\documentclass{article}
\usepackage{spconf,amsmath,graphicx}
\usepackage{booktabs} % 提供更好的表格线条
\usepackage{amssymb}
\usepackage{makecell} % 支持 \makecell 命令
\usepackage{multirow}
\usepackage{url}
\usepackage{algorithm}
 \usepackage{hyperref}
\title{Visual Saliency Steering Distillation for Multimodal Chain-of-Thought Reasoning}
\name{Hao Yang, Jin Wang\textsuperscript{\dag}\thanks{\textsuperscript{\dag} Corresponding author.}, Xuejie Zhang}
\address{School of Information Science and Engineering, Yunnan University, Kunming, China \\
  yanghao888@stu.ynu.edu.cn, \{wangjin, xjzhang\}@ynu.edu.cn}

\begin{document}
%\ninept
%

\maketitle
% \footnotetext{\textsuperscript{\dag}~Corresponding author.}

\begin{abstract}
Multimodal chain-of-thought (CoT) reasoning integrates visual and textual cues through step-by-step inference. In small models with limited token budgets, modality-interaction fusion often suppresses tiny cross-modal differences. In particular, multimodal CoT often struggles when different images pair with identical text or different texts pair with an identical image, making such inputs nearly indistinguishable after fusion. This study proposes Visual Saliency Steering Distillation (VSSD). VSSD leverages the attention maps of multimodal large language models to generate perturbed images that capture task-sensitive feature directions, and then applies singular value decomposition to extract dominant steering vectors to guide inter-layer distillation. Experiments on ScienceQA and M$^3$CoT demonstrate that VSSD improves rationale generation and answer inference. The code is available at \href{https://github.com/BGWH123/VSSD}{%
https://github.com/BGWH123/VSSD
}.

\end{abstract}


\section{Introduction}
% MCoT背景 -> MLLMs在MCoT上的方法，缺陷（成本昂贵，所以要用小模型） -> 小模型上的做法，缺陷 -> 分析原因，对视觉模态关注不足，大小模型都一样，大模型可引用multimodal steer减少幻觉的相关论文，小模型受限于模型规模会进一步加剧这个问题，例如特征融合 -> 提出的办法 

Multimodal chain-of-thought (CoT) captures visual-linguistic cues through explicit reasoning steps. A natural approach is to leverage multimodal large language models (MLLMs) \cite{hurst2024gpt, liu2023visual} to enable CoT. However, the large parameter size and high computational costs limit the deployment in resource-constrained scenarios. Therefore, research on multimodal CoT for small models is crucial.

With limited token budgets, existing multimodal CoT methods for small models rely on modality-interaction fusion to integrate information \cite{zhang2023multimodal,tan2024boosting,Liu2025,yuan2023joint}. However, such fusion often leads to the confusion of tiny cross-modal features. Therefore, some approaches focus on modeling fine-grained differences during the CoT generation process \cite{zheng-etal-2024-enhancing}. Others observe that modalities disadvantaged in data or parameter size often struggle to capture fine details, thus seeking to enhance the visual representation \cite{wei2024enhancing}. Despite these advances, current models still lack sufficient sensitivity and understanding of fine-grained distinctions across different inputs.

\begin{figure}[t]
    \centering

    \includegraphics[width=1.0\linewidth]{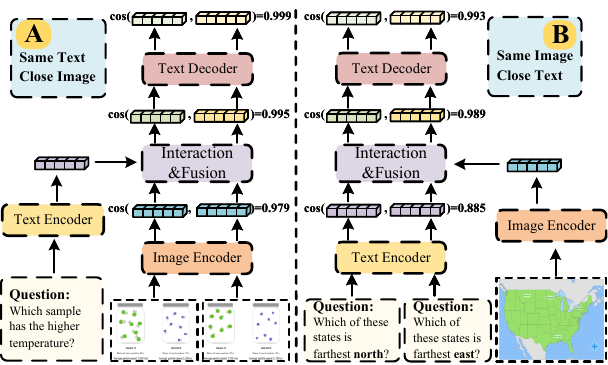}
    \caption{Fine-grained confusion in multimodal CoT.}
    \label{fig:vssd_difference}
\end{figure}

As illustrated in Fig.~\ref{fig:vssd_difference}, multimodal CoT faces two particularly challenging questions that significantly exacerbate the confusion of fine-grained differences: \textbf{(A)} different samples share identical text inputs but have similar image content, or \textbf{(B)} share identical image inputs but have similar textual descriptions. For unimodal, they can typically distinguish such inputs by relying on tiny cues within a single modality. For instance, even for similar inputs, features from individual modalities still retain discernible differences. However, for multimodal, the modality fusion process tends to suppress these discriminative signals further, making it difficult for models to capture critical distinctions. As shown in Fig.~\ref{fig:vssd_difference}, after cross-modal interaction and fusion, the feature representations of different inputs become significantly more similar and are nearly indistinguishable at the decoding stage.

This study proposes a visual saliency steering distillation (VSSD). Specifically, VSSD leverages MLLMs to identify and mask the image regions most relevant to the question, generating negatively perturbed semantic samples. By comparing the feature responses in the decoder between the original and the masked image, VSSD captures the sensitive direction critical to the task. Subsequently, singular value decomposition (SVD) is applied to extract the dominant steering vector from this difference, representing the optimal direction for amplifying fine-grained visual-semantic discrepancies. The steering vector is then injected into the intermediate layers of the model via layer-wise distillation, guiding the model to enhance its sensitivity to key visual-semantic differences. Extensive experiments on ScienceQA \cite{lu2022learn} and M$^3$CoT \cite{chen-etal-2024-m3cot} validate the effectiveness of VSSD.

\noindent\textbf{Our key contributions are as follows:} 
\textbf{\MakeUppercase{\romannumeral 1}.} We propose VSSD to address the limitation of small multimodal CoT models in capturing fine-grained visual-semantic differences. 
\textbf{\MakeUppercase{\romannumeral 2}.} VSSD leverages saliency-based perturbation and SVD-guided steering vector injection to enhance cross-modal sensitivity. 
\textbf{\MakeUppercase{\romannumeral 3}.} Experiments on ScienceQA and M$^3$CoT demonstrate that rational generation and answer inference consistently improved.

\section{Visual Saliency Steering Distillation}
\begin{figure}[t]
    \centering
    
    \includegraphics[width=1\linewidth]{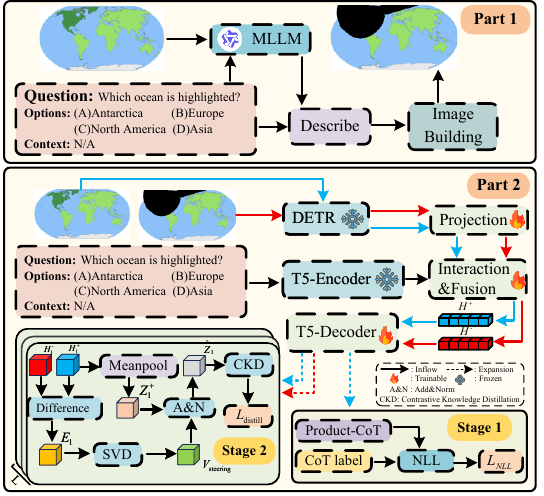}
    \caption{The overall framework of VSSD.}
    \label{fig:vssd}
\end{figure}

The proposed VSSD has two components in Fig.~\ref{fig:vssd}: (i) generating  perturbed image to emphasize key visual cues, and (ii) extracting the main semantic shift between original and perturbation features to guide inter-layer distillation.

\subsection{Preliminaries}

\noindent\textbf{Model Architecture}. 
We adopt a two-stage scheme: rationale generation and answer prediction. Given an image $v$ and textual input $x = Q \circ C \circ M$, where $Q$, $C$, $M$ denote the question, context, and choices respectively, the model first generates $r$. Then, $x$ and $r$ are concatenated and passed for answer prediction:
\begin{equation}
r = f(x, v), \quad
a = g(x \circ r, v),
\end{equation}

\noindent where $\circ$ denotes concatenation. $f$ and $g$ share architecture (T5) but are trained separately, with a frozen visual encoder to encourage language adaptation.

\noindent\textbf{Cross-Modal Fusion}.
Encoded image and text are $H_v$ and $H_l$ from DETR and T5. Cross-attention aligns them:
\begin{equation}
\hat{H}_v = \text{softmax} \left( \frac{(W_Q H_l)(W_K H_v)^\top}{\sqrt{d}} \right)(W_V H_v).
\end{equation}

Then, a gated fusion:
\begin{equation}
\sigma = \text{sigmoid}(W_l H_l + W_v \hat{H}_v), \quad
H_{\text{Enc}} = (1 - \sigma) \cdot H_l + \sigma \cdot \hat{H}_v.
\end{equation}
\noindent where $H_{\text{Enc}}$ is input to the decoder.

\noindent\textbf{Training and Inference}. 
Given target $y \in \{r, a\}$, the model is trained by minimizing the standard negative log-likelihood:
\begin{equation}
\mathcal{L}_{\text{NLL}} = -\sum_{i=1}^{N} \log p_\theta(y_i \mid x, v, y_{<i}).
\end{equation}

In inference, greedy decoding is applied in two stages, corresponding to different instantiations of $y$:
\begin{equation}
r = \arg\max_{r'} \, p(r' \mid x, v), \quad 
a = \arg\max_{a'} \, p(a' \mid r, x, v).
\end{equation}

\subsection{Image Perturbation Construction}

To emphasize informative regions and suppress spurious visual cues, Perturbed images are constructed guided by cross-attention maps from MLLMs. The process is summarized in Algorithm~\ref{alg:image-perturb}.

\begin{algorithm}[t]
\caption{Image Perturbation Construction}
\label{alg:image-perturb}
\textbf{Input:} Image $I$, question $Q$, threshold $\tau$, top-$k$, enhancement factor $\alpha$, constant $\epsilon$, number of layers $L$, number of decoding tokens $T$.\\
\textbf{Output:} Perturbed image $I_r$.

1.\textbf{Attention Extraction:}
$A = \text{MLLM}(I, Q)$; re-feed $(I, Q, A)$ to get cross-attention maps $\mathcal{M}$.

2.\textbf{Aggregation \& Projection:}  
$\bar{\phi} = \frac{1}{LT} \sum_{\ell,t} \mathcal{M}_{\ell,t,:}$;  
$\mathcal{A} = \text{Upsample}(\text{Reshape}(\bar{\phi}))$.

3.\textbf{Normalization:}  
$\hat{\mathcal{A}} = \alpha \cdot \frac{\mathcal{A} - \min(\mathcal{A})}{\max(\mathcal{A}) - \min(\mathcal{A}) + \epsilon}$;  
apply Gaussian blur $\to \mathcal{A}_{\text{smooth}}$.

4.\textbf{Mask Generation:}  
$M = \mathbb{1}[i \in \text{Top-k}] \cdot \mathbb{1}[\mathcal{A}_{\text{smooth}}(i) > \tau]$.

5.\textbf{Perturbation:}  
$I_r = I \odot (1 - M)$.

\end{algorithm}

First, cross-attention maps are extracted from the multimodal MLLM by re-feeding the image, question, and predicted answer. These maps are aggregated across layer and token dimensions to obtain a global visual importance vector, which is then reshaped and upsampled into a 2D spatial map. After normalization and Gaussian smoothing, a binary mask is generated using top-$k$ selection and thresholding. Finally, high-attention regions are masked from the original image, producing a perturbed version that preserves only less informative visual content.

\begin{table}[t]
\centering
\caption{Comparative results on ScienceQA. Size = backbone model size. Question classes: NAT = natural science, SOC = social science, LAN = language science, TXT = text context, IMG = image context, NO = no context, G1-6 = grades 1-6, G7-12 = grades 7-12. Results in bold are the best performance.}
\renewcommand{\arraystretch}{1.1}
\setlength{\tabcolsep}{1.2pt}
%\tiny
\fontsize{7pt}{8pt}\selectfont
\begin{tabular}{l|l|ccccccccc}
\hline
\textbf{Model} & \textbf{Size} & NAT & SOC & LAN & TXT & IMG & NO & G1-6 & G7-12 &  Avg \\
\hline
Human & - & 90.23 & 84.97 & 87.48 & 89.60 & 87.50 & 88.10 & 91.59 & 82.42 & 88.40 \\
\hline
\multicolumn{11}{l}{\textit{\textbf{LLM}}} \\
GPT-3.5 & 173B & 74.64 & 69.74 & 76.00 & 74.44 & 67.28 & 77.42 & 76.80 & 68.89 & 73.97 \\
GPT-3.5+CoT & 173B & 75.44 & 70.87 & 78.09 & 74.68 & 67.43 & 79.93 & 78.23 & 69.68 & 75.17 \\
ChatGPT+CoT & - & 78.82 & 70.98 & 83.18 & 77.37 & 67.92 & 86.13 & 80.72 & 74.03 & 78.31 \\
GPT-4+CoT & - & 85.48 & 72.44 & 90.27 & 82.65 & 71.49 & 92.89 & 86.66 & 79.04 & 83.99 \\
\hline
\multicolumn{11}{l}{\textit{\textbf{Finetune (VLLM)}}} \\
LLaMA-Adapter & 6B & 84.37 & 88.30 & 84.36 & 83.72 & 80.32 & 86.90 & 85.83 & 84.05 & 85.19 \\
SciTune\textsubscript{Base} & 7B & 84.50 & 94.15 & 82.91 & 88.35 & 83.64 & 88.74 & 85.05 & 85.60 & 86.11 \\
LaVIN & 13B & 90.32 & 94.38 & 87.73 & 89.44 & 87.65 & 90.31 & 91.19 & 89.26 & 90.50 \\
LLaVa & 13B & 90.36 & 95.95 & 88.00 & 89.49 & 88.00 & 90.66 & 90.93 & 90.90 & 90.92 \\
LLaVa (G4) & 13B & 91.56 & \textbf{96.74} & \textbf{91.09} & 90.62 & 88.99 & \textbf{93.52} & 92.73 & 92.16 & 92.53 \\
SciTune\textsubscript{Large} & 13B & 89.30 & 95.61 & 87.00 & 93.08 & 86.67 & 91.75 & 84.37 & 91.30 & 90.03 \\
\hline
\multicolumn{11}{l}{\textit{\textbf{Finetune (VSLM)}}} \\
Enigma-COT & 229M & 88.28 & 78.74 & 85.64 & 88.51 & 84.28 & 86.90 & 85.43 & 85.89 & 85.59 \\
MM-CoT\textsubscript{Base} & 223M & 87.52 & 77.17 & 85.82 & 87.88 & 82.90 & 86.83 & 84.65 & 85.37 & 84.91 \\
MM-CoT\textsubscript{Large} & 738M & \textbf{95.91} & 82.00 & 90.82 & 95.26 & 88.80 & 92.89 & 92.44 & 90.31 & 91.68 \\
DDCoT(T5) & 223M & 88.72 & 86.84 & 84.91 & 87.59 & 83.34 & 88.08 & 88.58 & 85.10 & 87.34 \\
VSSD\textsubscript{Base} & 223M & 93.61 & 79.98 & 89.45 & 93.84 & 86.47 & 90.94 & 90.60 & 88.00 & 89.67 \\
VSSD\textsubscript{Large} & 738M & 95.74 & 91.00 & 90.55 & \textbf{95.80} & \textbf{92.96} & 91.99 & \textbf{93.91} & \textbf{92.49} & \textbf{93.40} \\

\hline
\end{tabular}

\label{tab:scienceqa_main}
\end{table}

\subsection{Implicit Steering and Training Strategy}
After applying a gated fusion mechanism between the input image and the question, the decoding stage is performed. The final $L$ layers of hidden representations are extracted from both the original and perturbed images, denoted as $\{H^+_l\}_{l=1}^{L}$ and $\{H^-_l\}_{l=1}^{L}$, where $H^+_l, H^-_l \in \mathbb{R}^{B \times T \times D}$, and $B$, $T$, and $D$ denote the batch size, sequence length, and hidden dimension, respectively.

For each layer $l$, the difference tensor is computed, and SVD is performed:
\begin{equation}
E_l = H^+_l - H^-_l, \quad E^{(b)}_l = U \Sigma V^\top, \quad \forall b \in [1, B].
\end{equation}

Select the top-1 right singular vector as the principal editing direction:
\begin{equation}
\mathbf{v}_{\text{steering}}^{(b)} = V_{:,1}, \quad\mathbf{z}^+_l = \text{MeanPool}(H^+_l). 
\end{equation}

Collecting these steering vectors captures dominant semantic shifts caused by counterfactual visual information.

To leverage them, a layer-wise distillation mechanism is introduced. For each of the last $L$ decoder layers, mean pooling is applied over the token dimension to obtain compact hidden representations, and $\alpha$ is a scaling factor.

To ensure that the steering does not merely scale the hidden states but preserves their semantic direction, the steered representation is normalized to match the magnitude of the original representation:
\begin{equation}
\hat{\mathbf{z}}_l = \mathbf{z}^+_l + \alpha \cdot \mathbf{v}_{\text{steering}}, \quad
\tilde{\mathbf{z}}_l = \hat{\mathbf{z}}_l \cdot \frac{\|\mathbf{z}^+_l\|_2}{\|\hat{\mathbf{z}}_l\|_2 + \epsilon},
\end{equation}
where $\epsilon$ is a small constant.

The distillation loss and the overall training objective are defined as:
\begin{equation}
\mathcal{L}_{\text{distill}} = \frac{1}{L} \sum_{l=1}^{L} \big\|\mathbf{z}^+_l - \tilde{\mathbf{z}}_l\big\|_2^2, \quad
\mathcal{L} = \mathcal{L}_{\text{NLL}} + \beta \cdot \mathcal{L}_{\text{distill}}.
\end{equation}
where $\beta$ controls the trade-off between the main task loss and the steering-guided distillation loss.

%\label{sec:format}

\section{Experiment}
\subsection{Experiments Settings}

\textbf{Dataset}. The proposed VSSD was evaluated on the ScienceQA~\cite{lu2022learn} benchmark, a multimodal chain-of-thought dataset with over 21,000 multiple-choice questions across three science subjects. Further tests were conducted on M$^3$CoT~\cite{chen-etal-2024-m3cot}, a more challenging variant of ScienceQA where each sample is paired with an image.

\noindent\textbf{Implementation Details.} The T5 model is initialized with UnifiedQA \cite{khashabi2020unifiedqa}. The MLLMs are Qwen2.5-VL. Fine-tuning was conducted for up to 20 epochs with a learning rate 5e-5. The input lengths were 512 (rationale) and 64 (answer). Hyperparameters $\alpha$ and $\beta$  were 0.1 and 0.2. The layer $L$ is 2. A fixed seed 42 ensured reproducibility. 
% All experiments were run on a single NVIDIA A100 GPU.

\noindent\textbf{Baselines.} We compare our model with five categories: 
(1) \emph{Instruction-tuned LLMs}: GPT-3.5, CoT-enhanced variants, ChatGPT, GPT-4 \cite{lu2022learn}; 
(2) \emph{Tool-augmented LLMs}: Chameleon \cite{lu2023chameleon}, VisualChatGPT \cite{wu2023visual}, IdealGPT \cite{you2023idealgpt}; 
(3) \emph{Fine-tuned VLLMs}: LLaMA-Adapter \cite{zhang2023llama}, LaVIN \cite{luo2023cheap}, LLaVA \cite{liu2023visual}; 
(4) \emph{Visual SLMs}: MM-CoT \cite{zhang2023multimodal}, MC-CoT \cite{tan2024boosting} DDCoT \cite{zheng2023ddcot}, Enigma-COT \cite{wei2024enhancing}, and our VSSD.

\subsection{ Comparative Results}

\begin{table}[!t]
    \centering
    \caption{Comparative results on M$^3$CoT}
    \label{tab:m3cot_main}
    \setlength{\tabcolsep}{1pt} % 列间距
    \fontsize{6pt}{7pt}\selectfont
    \resizebox{0.45\textwidth}{!}{
    \begin{tabular}{lccccccc}
        \toprule
        \multirow{2}{*}{Model} &
        \multicolumn{2}{c}{Science} &
        \multicolumn{2}{c}{Commonsense} &
        \multicolumn{2}{c}{Mathematics} &
        \multirow{2}{*}{Total} \\
        \cmidrule(lr){2-3} \cmidrule(lr){4-5} \cmidrule(lr){6-7}
        & Lang & Natural & Physical & Temporal & Algebra & Geometry & \\
        \midrule

        \multicolumn{8}{l}{\textbf{Human \& Random}} \\
        Human & 97.83 & 92.62 & 96.28 & 88.71 & 87.23 & 88.75 & 91.61 \\
        Random & 32.70 & 30.62 & 32.97 & 20.33 & 35.71 & 27.50 & 28.56 \\
        \midrule

        \multicolumn{8}{l}{\textbf{Tool-Usage}} \\
        VisualChatGPT & 30.09 & 36.28 & 43.48 & 33.33 & 21.99 & 21.25 & 25.92 \\
        IdealGPT & 31.73 & 31.63 & 56.52 & 26.83 & 20.57 & 30.00 & 32.19 \\
        Chameleon & 43.87 & 26.05 & 39.13 & 48.78 & 17.73 & 26.25 & 34.29 \\
        \midrule

        \multicolumn{8}{l}{\textbf{Finetuning (VLLM)}} \\
        LLama-Adapter-7B & 62.56 & 72.29 & 76.92 & 72.36 & 30.71 & 38.75 & 54.89 \\
        LLaVA-v1.5-13B & 68.72 & 72.41 & 83.52 & 69.11 & 35.71 & 45.00 & 59.50 \\
        CogVLM-17B & 65.88 & 77.52 & 81.32 & 75.61 & 35.71 & 46.25 & 58.25 \\
        GPT4V w/CoT & \textbf{90.52} & 63.09 & \textbf{83.33} & 82.93 & 45.71 & 50.00 & 62.60 \\
        \midrule

        \multicolumn{8}{l}{\textbf{Finetuning (VSLM)}} \\
        MM-CoT-Large & 45.50 & 50.19 & 63.74 & 33.33 & 40.71 & 61.25 & 48.73 \\
        MC-CoT-Large & 42.65 & 67.43 & 58.24 & 56.10 & 57.86 & \textbf{62.50} & 57.69 \\
        VSSD-Base & 57.89 & \textbf{74.84} & 79.44 & \textbf{93.97} & \textbf{74.23} & 57.50 & \textbf{73.19} \\
        w/o distillation & 51.24 & 61.29 & 68.89 & 78.86 & 63.75 & 52.57 & 61.57 \\
        \bottomrule
    \end{tabular}
    }
    \label{tab:m3cot_main}
\end{table}

The ScienceQA results are reported in Table~\ref{tab:scienceqa_main}. VSSD\textsubscript{Base} (223M) achieves 89.67\% accuracy, outperforming Enigma-COT, MM-CoT\textsubscript{Base}, and DDCoT of similar size. VSSD\textsubscript{Large} (738M) further improves to 93.40\%, surpassing all other finetuned models, including the GPT-4-based LLaVA (92.53\%). The results on M$^3$CoT are shown in Table~\ref{tab:m3cot_main}, VSSD-Base reaches 73.19\% average accuracy, outperforming all fine-tuned baselines. Without inter-layer distillation, performance drops to 61.57\%, highlighting its importance for handling fine-grained multimodal reasoning.

\subsection{Perturbed Image Analysis}
Table~\ref{tab:two_stage_setting} compares the perturbed image (PI) with the black image (BKG) baseline. Models using PI consistently perform better in rationale generation and answer inference, as the perturbed image highlights critical visual regions and guides the model’s attention. In contrast, the BKG baseline removes informative regions, leading to a weaker focus on key evidence. This confirms the benefit of using perturbed images.

\begin{table}[t]
    \centering
    \scriptsize
    \setlength{\tabcolsep}{1.5pt} % 更紧凑的列间距
    \caption{Two-stage setting on ScienceQA: (i) rationale generation (RougeL), (ii) answer inference (Accuracy).}
    \label{tab:two_stage_setting}
    \begin{tabular}{@{}lcccc@{}} % 去掉表格左右额外间距
        \toprule
        & VSSD\textsubscript{Base}(PI) & VSSD\textsubscript{Base}(BKG) & VSSD\textsubscript{Large}(PI) & VSSD\textsubscript{Large}(BKG) \\
        \midrule
        QCM$\rightarrow$R (RougeL) & 98.11 & 97.81 & 98.43 & 98.13 \\
        QCMR$\rightarrow$A (Acc) & 89.67 & 87.95 & 93.40 & 90.64 \\
        \bottomrule
    \end{tabular}
\end{table}

\subsection{Steering Vector Layers and Information Analysis}
\begin{figure}[t]
\centering
\begin{minipage}[b]{0.51\linewidth}
  \centering
  \includegraphics[width=\linewidth]{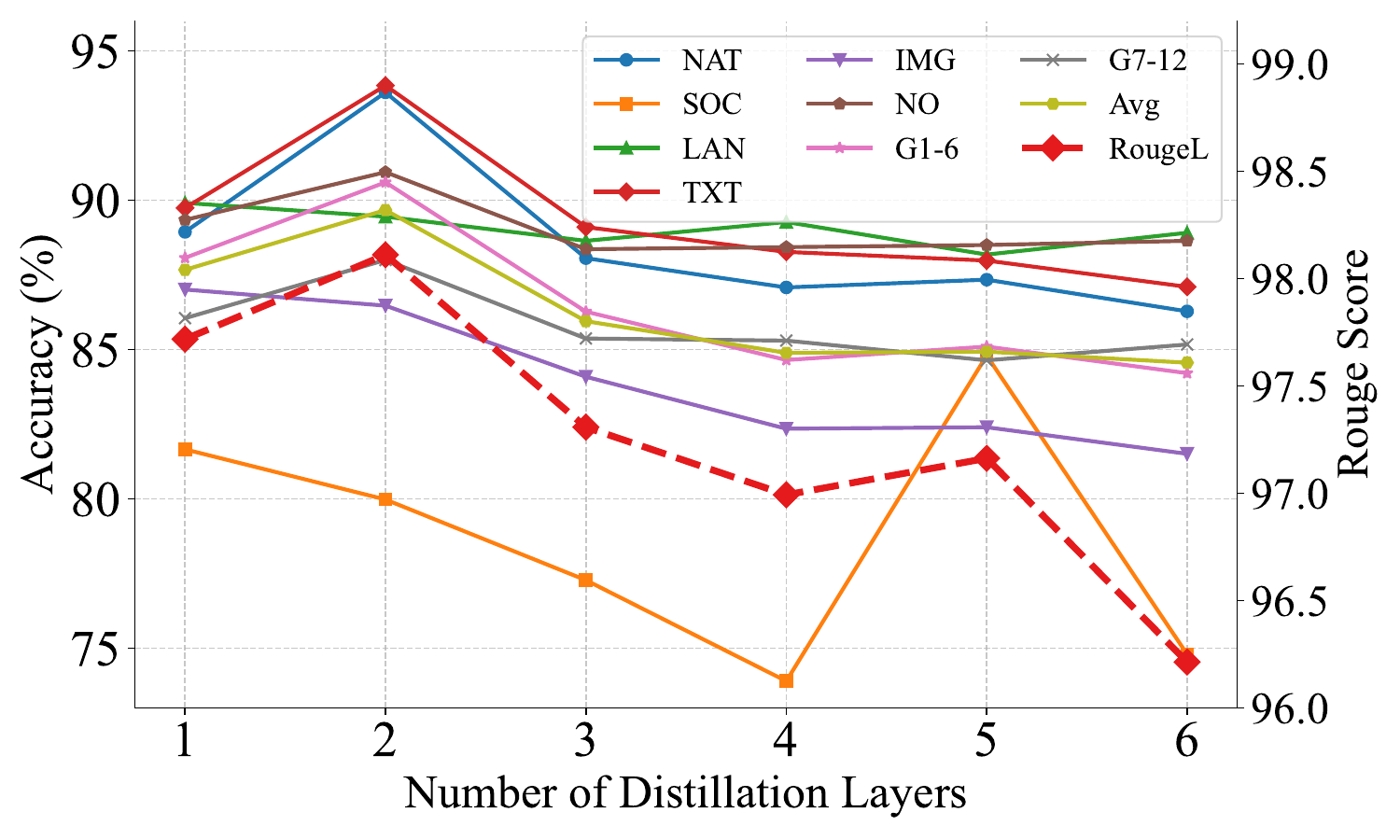}
  \centerline{(a) Performance vs Layer}\medskip
\end{minipage}
\begin{minipage}[b]{0.46\linewidth}
  \centering
  \includegraphics[width=\linewidth]{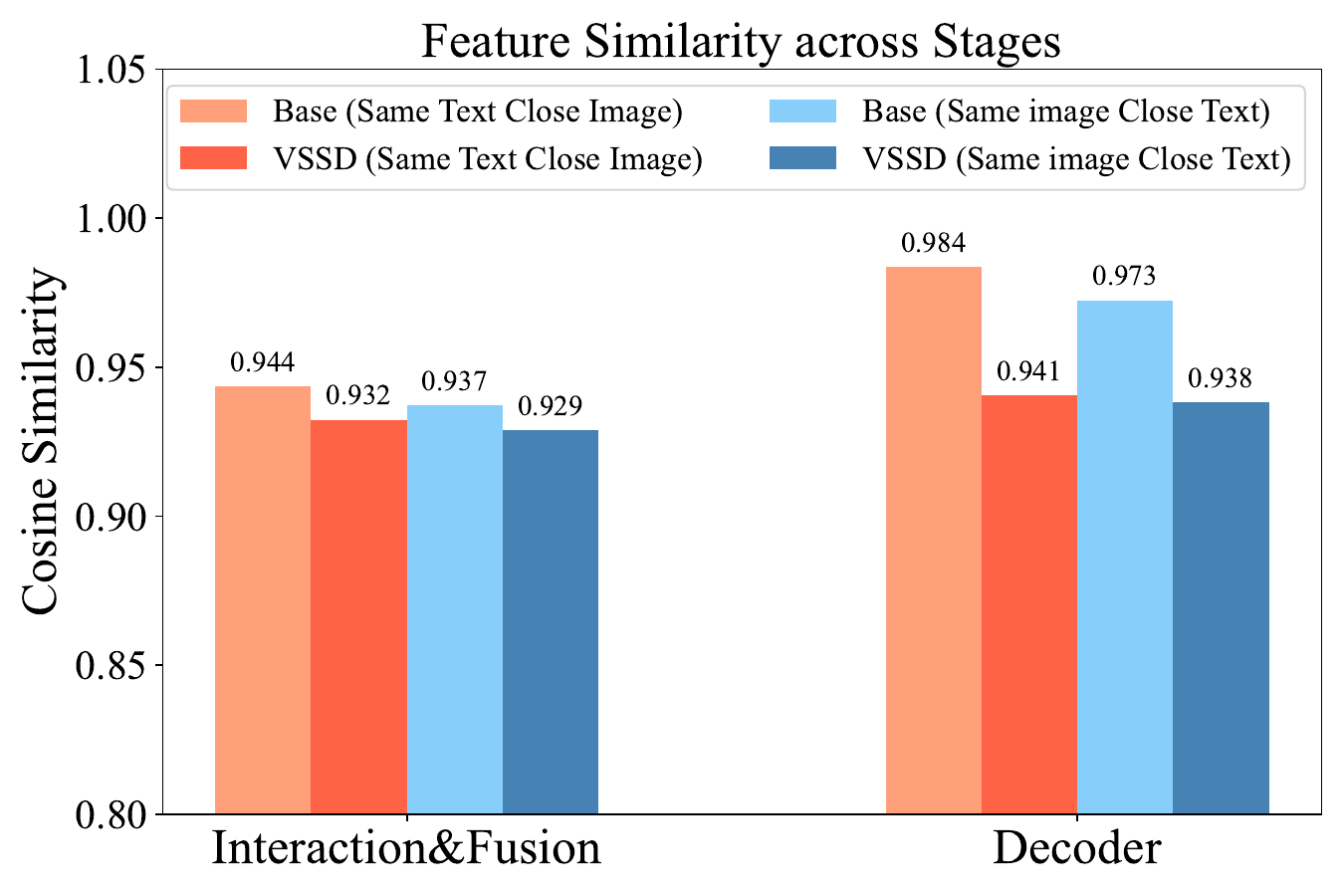}
  \centerline{(b) Cosine Similarity In Stages}\medskip
\end{minipage}

\hfill

\caption{Retained information and performance of VSSD on ScienceQA.}
\label{fig:VSSD_anl}

\end{figure}
A clear trend is revealed by the experimental results in Fig.~\ref{fig:VSSD_anl} (a). Using steering vectors from only the last $L=2$ decoder layers yields the best performance. For example, accuracy and RougeL scores reach their peaks at this setting. When fewer than two layers are used ($L=1$), the steering signal becomes too weak, as a single decoder layer cannot fully capture the semantic refinements necessary for reliable reasoning and generation. On the other hand, when more than two decoder layers are included ($L>2$), performance declines. The rationale is that earlier decoder layers mainly encode intermediate or low-level signals, which, when added, introduce redundant or noisy information that interferes with the decision-critical semantics of the final layers. Thus, extracting steering vectors from exactly the last two decoder layers provides the best balance—retaining sufficient high-level semantic shift while avoiding unnecessary noise.

% % Inspired by prior work on multimodal signal attribution \cite{akbari2019multi}, 
% To address the challenge of distinguishing visually or textually similar yet semantically distinct samples, we evaluate VSSD on two representative cases: 100 pairs of the same text close image and 100 pairs of the same image close text. As shown in Fig.~\ref{fig:VSSD_anl} (b), VSSD consistently reduces feature similarity in both cases across the Interaction \& Fusion and Decoder stages, indicating that it effectively suppresses superficial similarities and improves the model’s ability to capture subtle, meaningful differences.

% Fig.~\ref{fig:VSSD_anl}(a) shows a clear trend: using steering vectors from the last $L=2$ decoder layers yields the best performance, with accuracy and RougeL reaching their peaks. When only one layer is used ($L=1$), the steering signal is too weak because a single decoder layer cannot fully capture the semantic refinements needed for reliable reasoning and generation. When more than two layers are included ($L>2$), performance declines, since earlier decoder layers mainly encode intermediate or low-level signals that introduce redundant or noisy information. Therefore, extracting steering vectors from exactly the last two decoder layers provides the best balance between semantic strength and noise avoidance.

To address the challenge of distinguishing visually or textually similar yet semantically distinct samples, we evaluate VSSD on two representative cases: 100 pairs of the same text close image and 100 pairs of the same image close text. As shown in Fig.~\ref{fig:VSSD_anl} (b), VSSD consistently reduces feature similarity in both cases across the Interaction \& Fusion and Decoder stages, indicating that it effectively suppresses superficial similarities and improves the model’s ability to capture subtle, meaningful differences.
\subsection{Ablation Study}

\begin{table}[!t]
\centering
 
\setlength{\tabcolsep}{1.2pt}
\fontsize{7pt}{8pt}\selectfont
\caption{Ablation of VSSD\textsubscript{Base} on ScienceQA.}
\label{tab:ICDR_ablation}
\begin{tabular}{l|cccccccccc}
\hline
\textbf{Model} & \textbf{NAT} & \textbf{SOC} & \textbf{LAN} & \textbf{TXT} & \textbf{IMG} & \textbf{NO} & \textbf{G1-6} & \textbf{G7-12} & \textbf{Avg} & \textbf{R\textsubscript{RougeL}}  \\
\hline
VSSD\textsubscript{Base} & 93.61 & 79.98 & 89.45 & 93.84 & 86.47 & 90.94 & 90.60 & 88.00 & 89.67 & 98.11  \\
w/o PI                   & 89.34 & 81.55 & 88.36 & 89.69 & 86.91 & 89.27 & 88.03 & 86.42 & 87.46 & 97.74  \\
w/o ILD                  & 88.19 & 75.59 & 88.18 & 88.51 & 82.65 & 89.13 & 86.12 & 84.51 & 85.55 & 97.33  \\
\hline
\end{tabular}
\end{table}

An ablation study is conducted to investigate the contribution of key components in the framework. The results are presented in Table~\ref{tab:ICDR_ablation}.

% \noindent\textbf{Effect of perturbed Image(PI).}
% Removing the PI module (\textit{w/o PI}) leads to a noticeable performance drop across most categories, especially in NAT (–4.27), G1-6 (–2.57), and the overall average score (–2.21). This confirms that perturbed supervision helps the model better localize the visually relevant evidence by suppressing misleading attention areas. The performance on rationale generation (R\textsubscript{ROUGE}) and answer generation (A\textsubscript{ROUGE}) also declines, indicating that the contrastive input improves both s through enhanced reasoning focus.

% \noindent\textbf{Effect of Inter-Layer Distillation (ILD).}
% Disabling ILD (\textit{w/o ILD}) further deteriorates performance, with the lowest scores observed across all dimensions, including a 4.12 point drop in average accuracy compared to the full model. This suggests that ILD plays a crucial role in encoding subtle differences between factual and counterfactual inputs. By injecting steering vectors derived from SVD-guided latent differences, ILD enables the model to learn semantically meaningful variation across decoding layers. The reduced ROUGE scores also imply that without ILD, the model loses its sensitivity to structured visual changes, leading to degraded generation quality.
\noindent\textbf{Effect of perturbed Image (PI).}
Removing PI causes a noticeable performance drop across categories, confirming that perturbed supervision helps the model localize visually relevant evidence by suppressing misleading attention areas. The decline in rationale and answer generation indicates that contrastive input improves reasoning focus.

\noindent\textbf{Effect of Inter-Layer Distillation (ILD).}
Disabling ILD further deteriorates performance, showing its role in encoding tiny differences between factual and counterfactual inputs. ILD allows the model to capture semantically meaningful variations across decoding layers by injecting steering vectors derived from latent differences. The reduced generation quality suggests that the model is less sensitive to structured visual changes without ILD.

\section{Conclusion}
This study proposes a visual saliency steering distillation (VSSD) to address the loss of fine-grained visual-semantic differences in small multimodal CoT methods. Existing methods often blur tiny cross-modal cues. VSSD generates attention-guided perturbed image to capture task-sensitive feature directions and distills steering vectors into intermediate layers. Experiments on ScienceQA and M$^3$CoT show consistent improvements, validating the effectiveness of both components. Future work will extend VSSD to broader multimodal reasoning tasks and explore more adaptive steering vector extraction.

\section{Acknowledgement}
This work was supported in part by the National Natural Science Foundation of China (NSFC) under Grant Nos. 61966038 and 62266051, and the Postgraduate Research and Innovation Foundation of Yunnan University
under Grant No.KC-252513133. The authors would like to thank the anonymous reviewers for their constructive comments.

% Below is an example of how to insert images. Delete the ``\vspace'' line,
% uncomment the preceding line ``\centerline...'' and replace ``imageX.ps''
% with a suitable PostScript file name.
% -------------------------------------------------------------------------

%\vfill
%\pagebreak

\bibliographystyle{IEEEbib}
\bibliography{strings,refs}

\end{document}